\documentclass{article}

\usepackage[T1]{fontenc}
\usepackage{microtype}
\usepackage{graphicx}
\usepackage{subfigure}
\usepackage{amsmath}
\usepackage{amssymb}
\usepackage{booktabs} 
\usepackage{xcolor}

\usepackage{hyperref}

\usepackage[accepted]{icml2020}

\icmltitlerunning{On the Generalization Benefit of Noise in Stochastic Gradient Descent}

\begin{document}

\twocolumn[
\icmltitle{On the Generalization Benefit of Noise in Stochastic Gradient Descent}



\icmlsetsymbol{equal}{*}

\begin{icmlauthorlist}
\icmlauthor{Samuel L. Smith}{to}
\icmlauthor{Erich Elsen}{to}
\icmlauthor{Soham De}{to}

\end{icmlauthorlist}

\icmlaffiliation{to}{DeepMind, London}

\icmlcorrespondingauthor{Samuel L. Smith}{slsmith@google.com}
\icmlcorrespondingauthor{Soham De}{sohamde@google.com}

\icmlkeywords{SGD, momentum, batch size, learning rate, noise, temperature, implicit regularization, optimization, generalization}

\vskip 0.3in
]



\printAffiliationsAndNotice{}  

\begin{abstract}
It has long been argued that minibatch stochastic gradient descent can generalize better than large batch gradient descent in deep neural networks. However recent papers have questioned this claim, arguing that this effect is simply a consequence of suboptimal hyperparameter tuning or insufficient compute budgets when the batch size is large. In this paper, we perform carefully designed experiments and rigorous hyperparameter sweeps on a range of popular models, which verify that small or moderately large batch sizes can substantially outperform very large batches on the test set. This occurs even when both models are trained for the same number of iterations and large batches achieve smaller training losses. Our results confirm that the noise in stochastic gradients can enhance generalization. We study how the optimal learning rate schedule changes as the epoch budget grows, and we provide a theoretical account of our observations based on the stochastic differential equation perspective of SGD dynamics. 
\end{abstract}

\section{Introduction}

It has long been believed that stochastic gradient descent can generalize better than full batch gradient descent in deep learning \citep{heskes1993line, lecun2012efficient}. This topic was revived by \citet{keskar2016large}, who showed that the test accuracy often falls if one holds the learning rate constant and increases the batch size, even if one continues training until the loss ceases to fall. A number of recent papers have studied this effect \citep{smith2017bayesian, jastrzkebski2017three, chaudhari2018stochastic}. However this phenomenon has also been questioned by many authors \citep{hoffer2017train, shallue2018measuring, zhang2019algorithmic}. In a widely read work, \citet{shallue2018measuring} argue that much of the generalization benefit of small batches arises either because the learning rate is not properly tuned for large batches, or because authors often compare different batch sizes under a constant epoch budget (such that small batches are allowed to take more parameter steps). To our knowledge, no previous authors have observed a generalization gap between small and large batch training under a constant step budget after properly tuning the learning rate schedule.

This debate is particularly challenging to resolve, because there is no consensus regarding how SGD hyperparameters should be tuned. While large research labs can afford to run complete grid searches over multiple dimensions \cite{shallue2018measuring}, this option is unrealistic for most researchers. Many authors have proposed a linear scaling rule between learning rate and batch size \citep{krizhevsky2014one, goyal2017accurate, smith2017don, mccandlish2018empirical}, while others propose a square root rule \citep{hoffer2017train} or argue that no single scaling rule is reliable across multiple architectures \cite{shallue2018measuring}. Some authors argue that SGD with Momentum significantly outperforms vanilla SGD \cite{sutskever2013importance}, while others claim that SGD with and without Momentum are equivalent if one maintains a constant ``effective learning rate'' \cite{mandt2017stochastic, kidambi2018insufficiency, liu2018accelerating}. \citet{goyal2017accurate} found that learning rate warmup enables us to scale training efficiently to larger batch sizes, and \citet{shallue2018measuring} emphasized that the optimal scaling strategy may change, depending on whether one scales the batch size under a constant epoch budget or a constant step budget.

Fortuntely, recent theoretical work suggests a consensus may be within reach \citep{ma2017power, zhang2019algorithmic}. These papers clarify the debate, by observing that SGD has two regimes with different behaviours. We refer to these two regimes as the ``noise dominated'' regime, which arises when the batch size is small or the loss is well conditioned, and the ``curvature dominated'' regime, which arises when the batch size is large or the loss is poorly conditioned. Under certain assumptions, the linear scaling rule will hold in the noise dominated regime for constant epoch budgets \citep{ma2017power, smith2017bayesian, zhang2019algorithmic}. However this rule does not hold in the curvature dominated regime. Similarly, SGD with and without Momentum achieve similar performance in the noise dominated regime if one maintains a constant effective learning rate \citep{smith2019momentum, zhang2019algorithmic}, but SGD with Momentum performs better in the curvature dominated regime \cite{shallue2018measuring}. 

However, these works primarily consider convergence on the training set, often for convex losses. They do not resolve the debate regarding the role of stochastic gradients in promoting generalization. It is also not clear that hyper-parameter tuning strategies derived to optimize the convergence rate will apply if one wishes to maximize the validation accuracy. In this paper, we make progress on both of these open questions. We list our main contributions below.
\begin{itemize}
    \item We use the analogy between SGD and stochastic differential equations (SDEs) \citep{gardiner1985handbook, welling2011bayesian, mandt2017stochastic, li2017stochastic} to describe the noise dominated and curvature dominated regimes of SGD in section \ref{sec:two-regimes}. Although similar conclusions can be derived from convergence bounds \citep{ma2017power, zhang2019algorithmic}, the SDE perspective will help us make explicit predictions about the performance of SGD on the test set. We verify the existence of two regimes of SGD on a range of models under a constant epoch budget in section \ref{sec:constant-epoch} and appendix \ref{app:additional_results_constant_epoch}.
    
    \item We confirm empirically that small and moderately large batch sizes outperform very large batches on the test set in some models, even if all batch sizes are trained for the same number of iterations and large batches reach smaller training losses (see section \ref{sec:constant-step} and appendix \ref{app:additional_results_constant_step}). We perform a grid search over learning rates at each batch size. The batch size at which the test accuracy begins to degrade can be larger than previously thought. For example, we find that the test accuracy of a 16-4 Wide-ResNet \citep{zagoruyko2016wide} trained on CIFAR-10 for 9725 updates falls from 94.9\% at a batch size of 2048 to 92.5\% at a batch size of 16384. 
    
    \item We find that the optimal learning rates, which either minimize the training loss or maximize the test set accuracy, scale differently as the epoch budget rises. This effect is not captured by existing convergence bounds. Although the learning rate that minimizes the training loss falls rapidly as the epoch budget rises, the learning rate that maximizes the test set accuracy decays very slowly. For example, for the same 16-4 Wide-ResNet on CIFAR-10 at batch size 64, the optimal learning rate to maximize the test accuracy only decays by a factor of 2 when the epoch budget is increased by a factor of 128, while the optimal learning rate to minimize the training loss decays by a factor of 16 (see section \ref{sec:implicit-reg}). We give a simple explanation from the SDE perspective: SGD seeks to maintain an ``optimal temperature'' early in training, independent of compute budget. This maximizes the generalization benefit arising from gradient noise, and results in a large initial learning rate, even if the epoch budget is also large. We also explore optimizing the initial and final learning rates independently (see section \ref{subsec:double_sweep}).

\end{itemize}

\section{Preliminaries of Empirical Analysis of SGD}
\label{sec:prelim_budget_lr}

The $i^{th}$ update of minibatch gradient descent is given by
\begin{align}
\textstyle
\omega_{i+1} = \omega_i - \frac{\epsilon_i}{B} \sum_{j=1}^{B} \frac{dL(y_j,x_j,\omega_i)}{d\omega},
\label{sgd}
\end{align}
where $(x,y)$ denotes the inputs and labels of a training set of size $N$, $B$ is the batch size, $\epsilon_i$ is the learning rate used on the $i^{th}$ step, and $L(y_j,x_j,\omega)$ is the loss of the $j^{th}$ training example. For simplicity we assume the indices $j$ are randomly reshuffled between each update, such that training batches are sampled randomly without replacement. When $B=N$, we get the full batch gradient descent update. We denote the full batch loss by $C(\omega) = \frac{1}{N}\sum_{j=1}^N L(y_j,x_j,\omega)$.


It is clear from equation \ref{sgd} that the dynamics of SGD depend heavily on the learning rate schedule $\{\epsilon_i\}$ and the batch size $B$. In many of the experiments in this paper we will sweep over the batch size on a logarithmic grid, in order to understand the effect of noise in the gradient estimate on the final performance of models trained with SGD. To ensure our conclusions are robust, we have chosen a single simple learning rate decay schedule which performs well across all of the architectures and datasets considered in this work (see section \ref{sec:exp_setup}). This schedule is defined by a single free parameter, the initial learning rate (usually referred to simply as the \emph{learning rate}). We always perform a grid search over learning rates for each batch size.

Furthermore, the conclusions of any empirical SGD study will depend on the choice of compute budget used for the experiments \cite{shallue2018measuring}. There are three popular compute budgets often considered in previous work, shown below. We explore all three compute budgets in this work.

\begin{itemize}
    \item \emph{Constant epoch budget: } Here the computational cost is independent of the batch size, but the number of steps is inversely proportional to the batch size.
    
    \item \emph{Constant step budget: } Here the computational cost is proportional to the batch size, but the number of training steps is independent of the batch size.

    \item \emph{Unlimited compute budget: } Here we train for as long as needed to maximize the test accuracy, or until a predetermined threshold performance target is reached.

\end{itemize}

\section{A Stochastic Differential Equation Perspective on the Two Regimes of SGD}
\label{sec:two-regimes}

In this section, we discuss the \emph{noise dominated} and \emph{curvature dominated} regimes of SGD, from the perspective of the analogy between SGD and stochastic differential equations (SDEs) \citep{gardiner1985handbook, welling2011bayesian, mandt2017stochastic, li2017stochastic}.
Although the two regimes are also visible within existing convergence bounds \citep{ma2017power, zhang2019algorithmic}, the SDE perspective will help us make explicit predictions about the test set behaviour.


\subsection{Full batch gradients}


When training with full batch gradients, the learning rate that minimizes the training loss fastest is determined by the curvature of the loss function. 
To minimize this loss as quickly as possible,
we usually set the learning rate early in training as large as possible while avoiding divergences or instabilities. To build our intuition for this, we approximate the loss by a strictly convex quadratic, $C(\omega) \approx \frac{1}{2} \omega^\top H \omega$. Substituting this approximation into the gradient descent parameter update, we conclude that
$\omega_{i+1} = \omega_i - \epsilon  H \omega_i.$
In the eigenbasis of $H$, the updates are $\theta_{i+1} = \theta_i (I - \epsilon \Lambda)$. Here $\theta_i = V^\top \omega_i$, where $V$ is a matrix whose columns are the eigenvectors of $H$, $I$ is the identity matrix and $\Lambda$ denotes a diagonal matrix comprising the eigenvalues of $H$. 
The iterates will converge if the learning rate $\epsilon < \epsilon_{crit}$, where $\epsilon_{crit} = 2/\lambda_{max}$ is the critical learning rate, and $\lambda_{max}$ is the largest Hessian eigenvalue \cite{nesterov2013introductory}. We call this inequality the \textit{curvature constraint}, and the optimal initial learning rate with full batch gradients will be just below $\epsilon_{crit}$. Although the critical learning rate will perform poorly for high curvature directions of the loss, we can introduce learning rate decay to minimize the loss along these directions later in training \citep{ge2019step}. 
Of course, in realistic loss landscapes $\epsilon_{crit}$ may change during training.

Acceleration methods like Heavy-Ball Momentum (referred to as ``Momentum'' hereon) \citep{polyak1964some} were designed to enable faster convergence on poorly conditioned losses with full batch gradients. Momentum takes an exponential moving average of previous gradients, $ \omega_{i+1} = \omega_i - \epsilon \sum_{j=0}^i m^{i-j} \frac{dC}{d\omega}\big|_{w=w_i},$ where $m$ denotes the momentum coefficient. Gradients in high curvature directions, which often switch sign between updates, partially cancel out. This enables Momentum to take larger steps in low curvature directions while remaining stable in high curvature directions. This allows Momentum to minimize the training loss in fewer steps than full batch gradient descent \citep{goh2017momentum}. 

\subsection{Minibatch gradients}

In practice, we do not compute a full batch gradient, and instead estimate the gradient over a minibatch \citep{bottou2010large}. This introduces noise into our parameter updates. However when the batch size is large, and the number of training epochs is finite, the noise in the parameter updates is low, and therefore training is still governed by the curvature of the loss landscape (similar to full batch gradient descent). We call this large batch training regime \textbf{\emph{curvature dominated}}. When the batch size is in the curvature dominated regime, we expect the optimal initial learning rate to be determined by the critical learning rate $\epsilon_{crit}$.
On the other hand, when the batch size is small, we expect the optimal learning rate to be controlled by the noise in the parameter updates, and we call this training regime \textbf{\emph{noise dominated}}.

To build a model of the training dynamics in the noise dominated regime, we must make some assumptions. Following previous work \citep{mandt2017stochastic, li2017stochastic, smith2017bayesian, jastrzkebski2017three}, we assume the gradients of individual examples are independent samples from an underlying distribution, and that this distribution is not heavy tailed. When the training set size $N \gg B$ and the batch size $B \gg 1$, we can apply the central limit theorem to model the noise in a gradient update by a Gaussian noise source, whose covariance is inversely proportional to the batch size,
\begin{equation}
     \textstyle \left( \omega_{i+1} - \omega_i \right) \approx -\epsilon \Big( \frac{dC}{d\omega}\Big|_{\omega = \omega_i} + \frac{\nu_i}{\sqrt{B}} \Big).
\end{equation}
The noise source $\nu$ has mean $\mathbb{E}(\nu_i) = 0$ and covariance $\mathbb{E}(\nu_i \nu_j^\top) = F(\omega_i) \delta_{ij}$, where $F(\omega)$ is the empirical Fisher information matrix and $\delta_{ij}$ is the dirac delta function. We may now introduce the \emph{temperature} $T = \epsilon/B$ to obtain:
\begin{equation}
    \textstyle
    \left( \omega_{i+1} - \omega_i \right) \approx -\epsilon \frac{dC}{d\omega}\Big|_{\omega = \omega_i} + \sqrt{\epsilon T} \nu_i.
    \label{eq:sde}
\end{equation}
Equation \ref{eq:sde} describes the discretization of a stochastic differential equation (SDE) with step size $\epsilon$ and temperature $T$ \citep{gardiner1985handbook}. We expect the dynamics of SGD to follow the underlying SDE if the learning rate $\epsilon \ll \epsilon_{crit}$ and the assumptions above are satisfied. When equation \ref{eq:sde} holds and $\epsilon \ll \epsilon_{crit}$, any two training runs with the same temperature and the same epoch budget should achieve similar performance on both the training set and the test set (see appendix \ref{app:sde_derive} or \citet{li2017stochastic} for details). Consequently, we usually expect the learning rate to scale linearly with the batch size in the noise dominated regime. This was observed in many empirical studies \citep{krizhevsky2014one, goyal2017accurate, mccandlish2018empirical}. 
For completeness, we derive this linear scaling rule in appendix \ref{app:sde_derive}, where we show that it can be derived without assuming that the noise in a gradient update is Gaussian or that the batch size $B \gg 1$. This scaling rule also arises within an analysis of convergence rates on quadratic losses \citep{zhang2019algorithmic}. The linear scaling rule may not hold if the noise is long tailed or one of the other assumptions above is not satisfied \citep{shallue2018measuring, simsekli2019tail}. We give an example of a model that does not obey linear scaling in appendix \ref{app:additional_models_constant_epoch}. 

In the noise dominated regime, the optimal learning rate increases as the batch size rises, and therefore when the batch size rises, we will eventually invalidate the assumption $\epsilon \ll \epsilon_{crit}$ and enter the curvature dominated regime. There may be a transition phase at the boundary between the two regimes \citep{liu2018mass}, however one of the surprising conclusions from our experiments is that in practice this transition is often very sharp (see section \ref{sec:constant-epoch}). 

The gradients of individual examples are not independent if batch normalization is used \cite{ioffe2015batch}. The linear scaling rule will therefore hold only if the batch statistics are computed over a fixed number of training examples independent of the batch size. This scheme, known as ghost batch normalization \citep{hoffer2017train}, is often used by default when large batches are partitioned over multiple devices. In this work, we use ghost batch normalization in all experiments that include batch normalization layers.

\subsection{Consequences of the two regimes}
Many previous works have established that SGD with and without Momentum are equivalent in the small learning rate limit when $m$ is fixed \citep{orr1994momentum, qian1999momentum, yuan2016influence}. In this limit, the speed of convergence of SGD with Momentum is governed by the effective learning rate $\epsilon_{eff} = \epsilon/(1-m)$, and the temperature $T = \epsilon_{eff}/B$ \citep{mandt2017stochastic, smith2017bayesian}. We therefore expect SGD with and without Momentum to achieve the same final training losses and test accuracies in the noise dominated regime (where $\epsilon_{eff} \ll \epsilon_{crit}$), while SGD with Momentum should outperform vanilla SGD in the curvature dominated regime. This was previously observed by \citet{shallue2018measuring}. More generally, as proposed by \citet{zhang2019algorithmic}, \emph{we typically expect that any optimizer which was designed for faster optimization on poorly conditioned loss surfaces will only outperform SGD if the batch size is large enough}.


\citet{goyal2017accurate} introduced learning rate warmup, and found that it enabled stable training with larger batch sizes for some architectures/datasets. This procedure has a straightforward interpretation within the two regimes: \emph{if the critical learning rate increases early in training, then learning rate warmup will enable us to achieve larger learning rates without diverging at the start of training, which in turn enables efficient training with larger minibatches}.

\subsection{On learning rate schedules and compute budgets}

Note that, with a very carefully tuned learning rate schedule, many batch sizes might exhibit both the curvature dominated regime (typically early in training) and the noise dominated regime (late in training) \citep{sutskever2013importance, de2017automated, zhang2019algorithmic}. 
However it is usually not possible to identify schedules of this type within a realistic computation budget. Practitioners prefer simple learning rate schedules, often parameterized by an initial learning rate and a few sharp drops \citep{he2016deep}. These schedules are easy to tune, and they are also thought to generalize well \citep{smith2017don, li2019towards}. For these popular schedules, the optimal learning rate is generally determined by whether the initial phase of training is noise dominated or curvature dominated. We refer to entire training runs as being noise or curvature dominated for simplicity. Note that, in the noise dominated regime, these schedules are best thought of not as a sequence of learning rates, but rather as a sequence of temperatures, each of which are maintained for a given number of epochs \citep{smith2017don}. Just as we refer to the initial learning rate as \emph{the learning rate}, we often refer to the initial temperature as \emph{the temperature}.



Throughout this paper, we assume the compute budget is \emph{finite but reasonably large}. For very small compute budgets, training may be curvature dominated at all batch sizes \citep{mccandlish2018empirical}. Meanwhile, for infinitely large compute budgets, the noise in the gradients might dominate asymptotically, and therefore training may be noise dominated for any batch size $B < N$  \citep{sutskever2013importance}.


\subsection{The generalization benefit of noise}
The primary difference between convergence bounds and the SDE perspective of SGD arises when we consider whether SGD has a beneficial influence on generalization \citep{mandt2017stochastic, jastrzkebski2017three, park2019effect}. 
Convergence bounds on convex losses predict that we should always achieve smaller training losses if we increase the batch size and train for the same number of steps \citep{ma2017power, zhang2019algorithmic}. 
However if we believe that SGD noise can enhance generalization from train to test, then the test accuracy achieved may fall as the batch size rises.

According to the SDE perspective, the influence of gradient noise on training in the noise dominated regime is described by the temperature, while the parameters at the end of training are sampled from a probability distribution that depends on the temperature and the epoch budget. We therefore expect two training runs in the noise dominated regime to experience a similar generalization benefit from noise if their temperatures are equal. However if the batch size is large enough to enter the curvature dominated regime, we will not be able to maintain a constant temperature while keeping the learning  rate below the critical learning rate. This suggests that to verify whether SGD noise is beneficial for generalization, we should compare small batch training to very large batch sizes in the curvature dominated regime.


Furthermore, since the influence of gradient noise is described by the temperature, and since we argue that this noise plays an important role in generalization, we conjecture that that the optimal temperature that maximizes the test accuracy will be independent of the epoch budget. This implies that, for a fixed batch size, the optimal learning rate will not decay as the epoch budget increases. We emphasize that the benefits of noise primarily arise early in training (see section \ref{sec:implicit-reg}). Decaying the learning rate (temperature) later in training often substantially enhances the test set accuracy.

\section{Experimental Setup}
\label{sec:exp_setup}

\begin{figure*}[t]
\centering
\subfigure[]{\includegraphics[height=3.8cm]{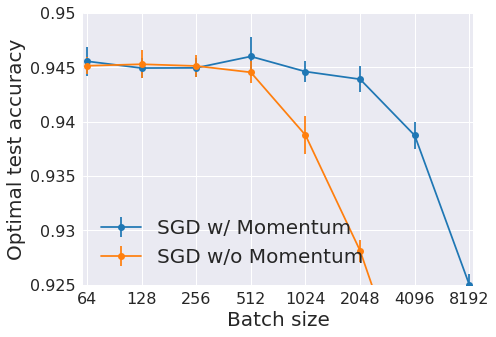}\label{fig:1a}}
\subfigure[]{\includegraphics[height=3.8cm]{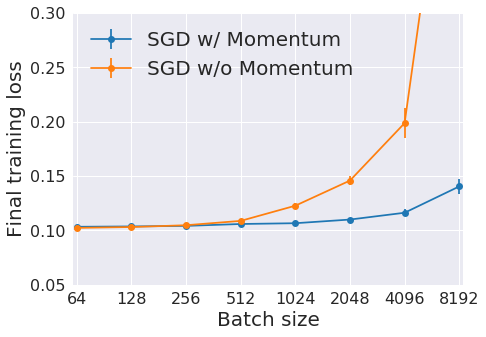}\label{fig:1b}}
\subfigure[]{\includegraphics[height=3.8cm]{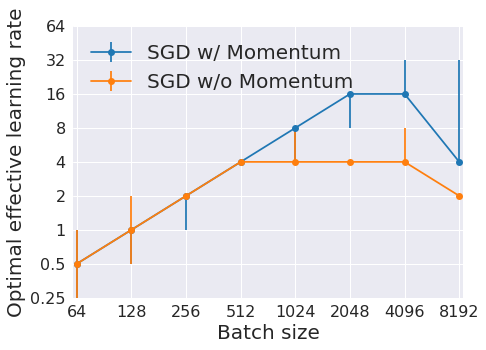}\label{fig:1c}}
\caption{A 16-4 Wide-ResNet, trained with ghost batch normalization on CIFAR-10 for 200 epochs. We report the performance of SGD with and without Momentum, and we perform a grid search to identify the optimal learning rate which maximizes the test set accuracy. a) The test accuracy of vanilla SGD is independent of batch size when the batch size is small, but falls when the batch size exceeds $512$. SGD with Momentum matches the performance of vanilla SGD for batch sizes $B \lesssim 512$ but outperforms vanilla SGD for batch sizes $B \gtrsim 512$. b) Very similar observations can be made for the final training loss. c) The optimal effective learning rate is proportional to batch size when the batch size is small, but is constant when the batch size is large. SGD with Momentum can scale to larger effective learning rates.}
\label{fig:two_regimes_constant_epoch_wrn}
\end{figure*}

In this paper, we will study how the performance on both the training and the test set, as well as how the optimal learning rate, depend on the batch size under different compute budgets (when using a realistic learning rate decay schedule). 
For clarity, in the main text we only report experiments using Wide-ResNets on CIFAR-10 \citep{zagoruyko2016wide}, however we provide additional experiments using ResNet-50 \citep{he2016deep}, LSTMs \citep{zaremba2014recurrent} and autoencoders \citep{sutskever2013importance} in the appendices. We describe the other models we study in appendix \ref{app:add_exp_details}. We use the same learning rate schedule for all architectures. 
We hold the learning rate constant for the first $N_{epochs}/2$ epochs, where $N_{epochs}$ denote the number of training epochs. Then for the remainder of training, we reduce the learning rate by a factor of $\gamma$ every $N_{epochs}/20$ epochs. In almost all of our experiments, we fix $\gamma = 2$, such that this scheme has a single hyperparameter, the initial learning rate $\epsilon$. We illustrate this schedule in appendix \ref{app:learning_rate_schedule}, and we found that it reliably meets or exceeds the performance of the schedules used by the authors of the original papers. We tune $\epsilon$ and $\gamma$ simultaneously in section \ref{subsec:double_sweep}.


We evaluate the optimal test accuracy and the optimal learning rate for a range of batch sizes and compute budgets. For each batch size, we train the Wide-ResNet model 15 times for a range of learning rates on a logarithmic grid. For each learning rate in this grid, we take the best 12 runs and evaluate the mean and standard deviation of their test accuracy. The optimal test accuracy is defined by the maximum value of this mean, and the corresponding learning rate is the optimal learning rate. This procedure ensures our results are not corrupted by outliers or failed training runs. To define error bars on the optimal learning rate, we include any learning rate whose mean accuracy was within one standard deviation of the mean accuracy of the optimal learning rate, and we always verify that both the optimal learning rate and the error bars are not at the boundary of our learning rate grid. We apply data augmentation including padding, random crops and left-right flips. The momentum coefficient $m=0.9$, the L2 regularization coefficient is $5\times 10^{-4}$, and when batch normalization is used we set the ghost batch size to 64 \citep{hoffer2017train}. We also report the mean final training loss at the optimal learning rate. We note that although we tune the learning rate on the test set, our goal in this paper is not to report state of the art performance, but rather to compare the performance at different batch sizes and with different training procedures. We apply the same experimental protocol in each case \citep{shallue2018measuring}. 

\section{SGD under a Constant Epoch Budget}
\label{sec:constant-epoch}


\begin{table*}[t]
\centering
\caption{The optimal test accuracy and final training loss for a range of batch sizes under a constant step budget. For each batch size, we train a 16-4 Wide-ResNet with ghost batch normalization for 9765 updates, and we perform a grid search to identify the optimal learning rate which maximizes the test set accuracy. The final training loss falls as the batch size increases, but the optimal test accuracy drops significantly for batch sizes greater than $2048$. This strongly supports that claim that minibatch gradient noise can enhance generalization. \\}
\begin{tabular}{c|c|c|c}
\textbf{Batch size} & \textbf{Optimal test accuracy (\%)} & \textbf{Final training loss} & \textbf{Optimal effective learning rate} \\ \hline
256                 & $93.5 \pm 0.1$                      & $0.232 \pm 0.001$            & $2^2$ ($2^1$ to $2^2$)                   \\
512                 & $94.2 \pm 0.1$                      & $0.171 \pm 0.001$            & $2^2$ ($2^2$ to $2^3$)                   \\
1024                & $94.5 \pm 0.1$                      & $0.107 \pm 0.001$            & $2^3$ ($2^3$ to $2^3$)                   \\
2048                & $94.9 \pm 0.1$                      & $0.058 \pm 0.000$            & $2^3$ ($2^3$ to $2^3$)                   \\
4096                & $94.7 \pm 0.1$                      & $0.025 \pm 0.000$            & $2^4$ ($2^4$ to $2^5$)                   \\
8192                & $94.6 \pm 0.1$                      & $0.017 \pm 0.000$            & $2^2$ ($2^2$ to $2^2$)                   \\
16384               & $92.5 \pm 0.6$                      & $0.019 \pm 0.004$            & $2^5$ ($2^4$ to $2^5$)                   \\
32768               & $89.9 \pm 0.7$                      & $0.039 \pm 0.011$            & $2^5$ ($2^0$ to $2^5$)                  
\end{tabular}
\label{table:1}
\end{table*}

In order to verify empirically that the two regimes of SGD arise on the test set as well as the training set, we perform a sweep over batch sizes under a fixed epoch budget. We train for the same number of epochs reported in the original papers, i.e., 200 epochs for Wide-ResNet on CIFAR-10.

In figure \ref{fig:1a}, we plot the optimal test accuracy for a range of batch sizes with a 16-4 Wide-ResNet, trained with batch normalization using SGD with and without Momentum. Both methods have the same optimal test accuracy when the batch size is small, but SGD with Momentum performs better when the batch size is large. The optimal test accuracy is independent of batch size when the batch size is small, but begins to falls when the batch size is sufficiently large. A similar trend is observed for the final training loss at the optimal effective learning rate in figure \ref{fig:1b}. To understand these results, we plot the optimal effective learning rate against batch size in figure \ref{fig:1c} (for SGD, $\epsilon_{eff} = \epsilon$). For SGD without Momentum, the learning rate is proportional to the batch size below $B \approx 512$, beyond which the optimal learning rate is constant. SGD with Momentum has the same optimal effective learning rate in the small batch limit, but it is able to scale to larger effective learning rates when $B \gtrsim 512$. All of these results exactly match theoretical predictions based on convergence bounds \citep{ma2017power, zhang2019algorithmic} or the SDE analogy (See section \ref{sec:two-regimes}).

The behaviour of SGD is strongly influenced by batch normalization \citep{bjorck2018understanding, santurkar2018does, sankararaman2019impact, park2019effect}. We therefore repeat this experiment without normalization in appendix \ref{app:without_BN_constant_epoch}. To ensure training is stable without batch normalization we use the ``Regularized SkipInit'' initialization scheme \citep{de2020batch}. We provide the full results of a learning rate sweep at two batch sizes in appendix \ref{app:lr_sweep_wrn}, as well as similar experiments for a range of models in appendix \ref{app:additional_models_constant_epoch}. 

\section{SGD under a Constant Step Budget}
\label{sec:constant-step}

In the section above, we studied training under a constant epoch budget, and we saw that SGD transitions between two regimes with different behaviours in a range of popular architectures. 
However, the results of the previous section do not tell us whether small batch training/minibatch noise has a generalization benefit which enhances the test set accuracy, because under a constant epoch budget large batches perform worse on both the training set and the test set.

To establish whether minibatch noise enhances generalization, we now evaluate how the optimal test accuracy depends on the batch size under a \emph{constant step budget}. This scheme ensures that large batch sizes have equal opportunity to minimize the training loss. In table \ref{table:1}, we report the optimal test accuracy of the 16-4 Wide-ResNet on CIFAR-10 at batch sizes ranging from 256 to 32768. For each batch size, we train for 9765 updates using SGD with Momentum (this corresponds to 200 epochs when the batch size is 1024). Following our previous learning rate schedule, we hold the learning rate constant for 4882 updates, and then decay the learning rate by a factor of 2 every 488 steps. We find that the optimal test accuracy initially increases with increasing batch size, but it then begins to fall sharply. The optimal test accuracy at batch size 2048 is $94.9\%$, but the optimal test accuracy at batch size 16384 is just $92.5\%$. For completeness, we have verified that batch size 16384 does not achieve higher test accuracies with smaller step budgets. We also report the final training loss, which falls as the batch size increases, as one would expect from convergence bounds \citep{zhang2019algorithmic}. To our surprise, the final training loss did rise at the largest batch size considered of 32768 examples, however we note that the training loss at this batch size is still smaller than the training loss achieved at batch size 2048, despite the test accuracy being $5\%$ lower.

All experiments in table 1 use a ghost batch size of 64 \citep{hoffer2017train}, which ensures that the noise arising from estimating the batch statistics on a subset of the training set does not change when the batch size rises. \citet{hoffer2017train} and \citet{de2020batch} showed that the test accuracy degrades when the ghost batch size is too large. In appendix \ref{app:additional_results_constant_step}, we also observed a drop in the test accuracy for very large batch sizes (under constant step budgets) when training a 16-4 Wide-ResNet without batch normalization on CIFAR-10, a 28-10 Wide-ResNet with or without batch normalization on CIFAR-100, as well as the autoencoder and LSTM tasks. These results confirm that stochastic gradient noise can enhance generalization. Although this effect was observed previously \citep{keskar2016large, smith2017bayesian, jastrzkebski2017three}, our experiment is the first to confirm it when training a popular model with a properly tuned learning rate schedule and a fixed step budget.

\citet{shallue2018measuring} argued that small batch sizes perform better under constant epoch budgets and large batch sizes perform better under constant step budgets. Our results clarify this claim, demonstrating that the test accuracy under constant step budgets initially improves with batch size but may degrade for very large batches. We note that \citet{shallue2018measuring} already observed that large batch sizes perform worse on the test set for ResNet-50/ImageNet, providing further evidence for our claims. They argue the performance gap between small and large batch sizes in this setup can be reduced by introducing additional explicit regularization.


\begin{figure*}[t]
\centering
\subfigure[]{\includegraphics[height=3cm]{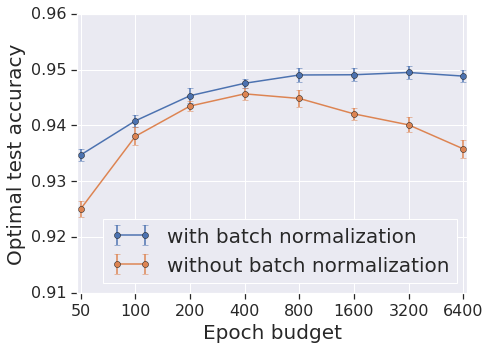}\label{fig:cifar_lr_vs_epoch_budget_a}}
\subfigure[]{\includegraphics[height=3cm]{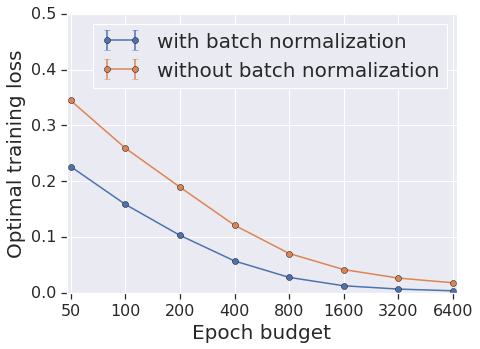}\label{fig:cifar_lr_vs_epoch_budget_b}}
\subfigure[]{\includegraphics[height=3cm]{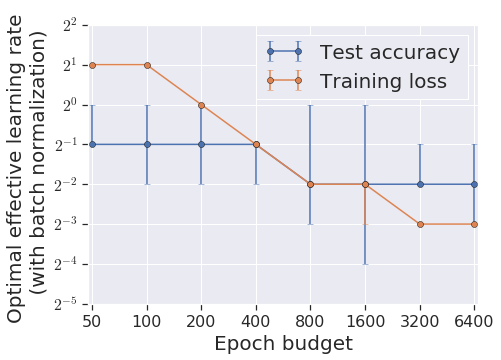}\label{fig:cifar_lr_vs_epoch_budget_c}}
\subfigure[]{\includegraphics[height=3cm]{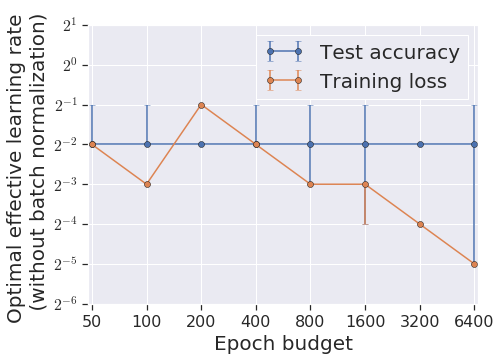}\label{fig:cifar_lr_vs_epoch_budget_d}}
\caption{A 16-4 Wide-ResNet trained on CIFAR-10 at batch size of 64 for a range of epoch budgets. We train with and without batch normalization. We identify both the optimal effective learning rate which maximizes the test accuracy and the optimal effective learning rate which minimizes the training loss. a) Initially the test accuracy rises as the epoch budget increases, however when training without batch normalization it begins to fall beyond $400$ training epochs, while with batch normalization it saturates after 800 epochs. b) The training loss falls monotonically as the epoch budget rises. c) With batch normalization, the learning rate which minimizes the training loss falls rapidly as the epoch budget rises, while the learning rate which maximizes the test accuracy only varies by a factor of 2 when the epoch budget rises over two orders of magnitude. d) Similarly, without batch normalization, the learning rate which minimizes the training loss falls as the epoch budget rises while the learning rate which maximizes the test accuracy is constant for all epoch budgets considered.}
\label{fig:cifar_lr_vs_epoch_budget}
\end{figure*}

\section{SGD with an Unlimited Epoch Budget}
\label{sec:implicit-reg}

We established in section \ref{sec:constant-step} that, in some popular architectures and datasets, the noise introduced by stochastic gradients does enhance generalization
. This motivates the following question: \emph{if the batch size is fixed, how does the optimal test accuracy and optimal learning rate depend on the epoch budget?} In particular, is the optimal training temperature ($T = \epsilon/B$) independent of the epoch budget, or does it fall as the number of training epochs increases?

To answer this question, we select a fixed batch size of 64, and we evaluate both the optimal test accuracy and the optimal training loss for a range of epoch budgets using SGD with Momentum. 
To study the effect of the optimal training temperature, we now \emph{independently measure both the optimal learning rate to maximize the test accuracy, and the optimal learning rate to minimize the training loss}. The optimal test accuracy and optimal training loss are shown in figures \ref{fig:cifar_lr_vs_epoch_budget_a} and \ref{fig:cifar_lr_vs_epoch_budget_b}. We train both with batch normalization and without batch normalization (using Regularized SkipInit \citep{de2020batch}), and we provide the optimal learning rates with batch normalization in figure \ref{fig:cifar_lr_vs_epoch_budget_c}, and the optimal learning rates without batch normalization in figure \ref{fig:cifar_lr_vs_epoch_budget_d}. 

In figure \ref{fig:cifar_lr_vs_epoch_budget_a}, we see that the optimal test accuracy initially increases, but then saturates or begins to fall as we increase the epoch budget further. This is similar to the well-known phenomenon of early stopping \citep{prechelt1998early, caruana2001overfitting}. Furthermore, in figure \ref{fig:cifar_lr_vs_epoch_budget_b}, we find that the optimal training loss falls monotonically as the epoch budget increases, consistent with classical optimization theory.

Figures \ref{fig:cifar_lr_vs_epoch_budget_c} and \ref{fig:cifar_lr_vs_epoch_budget_d} are more surprising. The learning rate that minimizes the training loss falls rapidly as the epoch budget rises. This is exactly what one would expect from convergence bounds on convex losses \citep{ma2017power, zhang2019algorithmic}. Strikingly however, when training with batch normalization, the learning rate that maximizes the test accuracy only falls by a factor of 2 when we increase the epoch budget from 50 to 6400 epochs. Meanwhile when training without batch normalization, the learning rate that maximizes the test accuracy is constant for all epoch budgets considered. These results support the claim that when training deep networks on classification tasks, there is an optimal temperature \citep{smith2017bayesian, park2019effect}, which biases small batch SGD towards parameters that perform well on the test set. 
We provide additional experimental results on other architectures in appendix \ref{app:additional_results_variable_epoch_budget}.



\subsection{Checking the robustness of our conclusions}
\label{subsec:double_sweep}

In the previous section, we show that there may be an optimal temperature during training that promotes good generalization performance. However the learning rate schedules used for these experiments have the property that the initial learning rate (denoted by say $\epsilon_0$) is coupled with the final learning rate (denoted by say $\epsilon_f$). More specifically, we have $\epsilon_f = \epsilon_0 \cdot \gamma^{-10}$, where $\gamma$ denotes the decay factor, which we set to 2 in our experiments (see appendix \ref{app:learning_rate_schedule}).

\begin{figure*}[t]
\centering
\subfigure[]{\includegraphics[height=2.9cm]{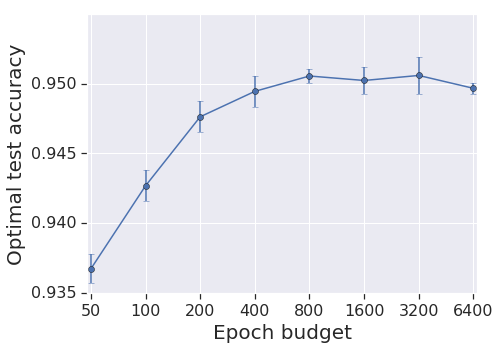}\label{fig:double_sweep_a}}
\subfigure[]{\includegraphics[height=2.9cm]{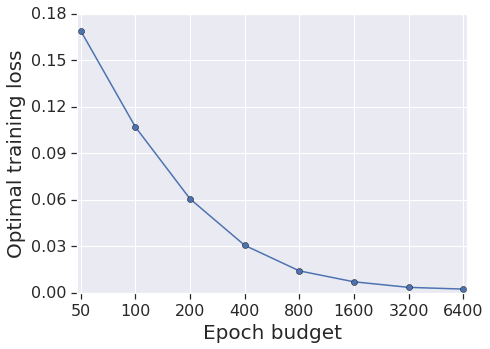}\label{fig:double_sweep_b}}
\subfigure[]{\includegraphics[height=2.9cm]{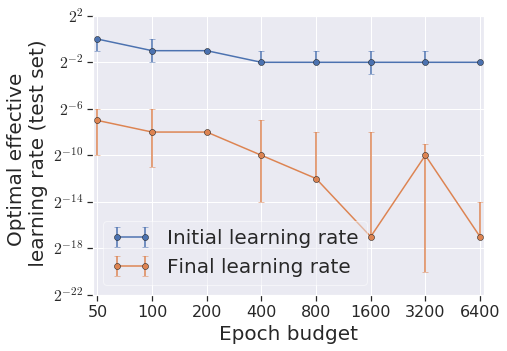}\label{fig:double_sweep_c}}
\subfigure[]{\includegraphics[height=2.9cm]{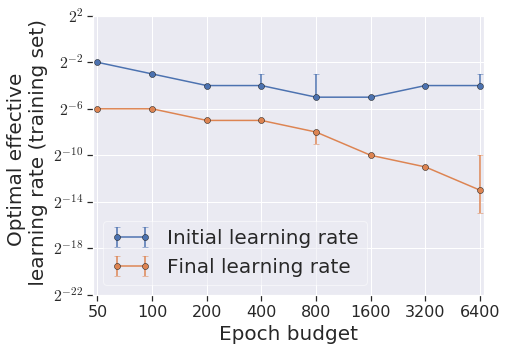}\label{fig:double_sweep_d}}
\caption{A 16-4 Wide-ResNet with batch normalization trained on CIFAR-10 at a batch size of 64 for a range of epoch budgets. We tune the initial and the final learning rates independently. We plot both the optimal initial and final learning rates for maximizing the test set accuracy, as well as the optimal initial and final learning rates for minimizing the training set loss. a) The test accuracy initially increases with increasing compute budget before saturating for epochs budgets greater than 800. b) Meanwhile the training loss falls monotonically as the epoch budget rises. c) The optimal initial learning rate which maximizes the test accuracy is constant for epoch budgets greater than 400, while the optimal final learning rate decays rapidly as the epoch budget increases. d) The optimal initial learning rate which minimizes the training loss decays slowly as the epoch budget increases, while the optimal final learning rate decays more rapidly.}
\label{fig:double_sweep}
\end{figure*}

Although common practice, coupling the initial and final learning rates makes it unclear whether the optimal temperature arises at the start or the end of training. It also does not optimize the decay factor. In figure \ref{fig:double_sweep}, we present the results of experiments with varying epoch budgets where we tune the initial and final learning rates independently. As in our previous experiments, when training for $N_{epochs}$ epochs, we use the initial learning rate for the first $N_{epochs}/2$ epochs, and then decay the learning rate by a factor of $\gamma$ every $N_{epochs}/20$ epochs. To define $\gamma$, we select an initial learning rate $\epsilon_0$ and a final learning rate $\epsilon_f$, and we then set $\gamma = (\epsilon_0/\epsilon_f)^{1/10}$. These experiments require a very large compute budget, and so we only study the 16-4 Wide-ResNet model with batch normalization at a batch size of 64 using SGD with Momentum. We consider epoch budgets between 50 and 6400 epochs, and we evaluate the optimal initial and final learning rates independently to both maximize the test accuracy and minimize the training loss. We evaluate the average performance of the best 5 out of 7 runs.


From figure \ref{fig:double_sweep}, we see that our main claims in the previous section still hold.
In addition, we make several observations from figures \ref{fig:double_sweep_c} and \ref{fig:double_sweep_d}.
We see that the optimal initial learning rate for maximizing the test set accuracy decays very slowly as the epoch budget rises, and it is constant for epoch budgets greater than 400. This supports the existence of an optimal temperature \emph{early in training} that boosts generalization performance. Meanwhile, the optimal final learning rate for maximizing the test set accuracy does decay rapidly as the epoch budget increases, which is likely helpful to prevent overfitting at late times. We note that the error bars on the final learning rate are much larger than those on the initial learning rate, suggesting that it is the initial learning rate which is most important to tune in practice. 

Furthermore, the optimal initial learning rate for maximizing the test accuracy is consistently higher than the optimal initial learning rate for minimizing the training loss, while the optimal final learning rate for maximizing the test accuracy is consistently lower than the optimal final learning rate for minimizing the training loss. These two observations support the widely held belief that learning rate schedules that maintain a high temperature at early times, and then decay the learning rate rapidly at late times, generalize well \citep{li2019towards}. There is a natural analogy between schedules of this type and simulated annealing \citep{smith2017don}. Rapidly decaying the temperature after an initial large learning rate phase ensures that the final parameters do not ``forget'' the influence of noise early in training.

\section{Discussion}
\label{sec:practical_recs}

In this paper, we study the generalization benefit of noise in stochastic gradient descent. We demonstrate that smaller batch sizes can outperform very large batch sizes on the test set under both constant epoch and constant step budgets, even after careful hyper-parameter tuning. Furthermore, when considering unlimited compute budgets, we find evidence of an ``optimal temperature'' that promotes generalization \citep{smith2017bayesian, park2019effect}. In most models, this temperature is defined by the ratio of the learning rate to the batch size early in training. Consequently, for a fixed batch size the existence of an optimal temperature implies that the optimal learning rate early in training will remain large even for very large compute budgets \citep{li2019towards}.

Although we have been careful in designing our experiments and in performing rigorous hyperparameter tuning, our conclusions are only valid for the learning rate schedule we used and the architectures we considered. We designed our schedule to ensure these conclusions are likely to apply to the popular schedules used by practitioners. However, given enough compute, it may be possible to design schedules that work equally well but do not follow our main claims. 

Our results suggest that, given a limited compute budget, one should save resources by choosing a schedule parameterized solely by the initial learning rate (e.g. fixing $\gamma = 2$ in our default schedule). One can also save resources by estimating the optimal learning rate on a logarithmic grid for a small epoch budget, before increasing the epoch budget to fine-tune. Given additional resources, one should also tune the final learning rate, but this usually has less influence on the test accuracy. If one wishes to reduce the wall clock time by parallelizing over large batches, a good rule of thumb is to train near the boundary between the noise and curvature dominated regimes \citep{mccandlish2018empirical}. One can estimate the location of this boundary by first running a cheap sweep over a few epochs to identify the largest stable learning rate, before scaling the batch size accordingly.


Despite a great deal of research, SGD with Momentum remains the most popular optimization algorithm in deep learning. Our research suggests two explanations for this. First, most optimization research designs algorithms for poorly conditioned losses. However typical batch sizes, $32 \lesssim B \lesssim 128$, are often in the noise dominated regime. In this regime, the training dynamics is governed by gradient noise, not conditioning. Algorithms designed to tackle curvature are more likely to help when the batch size is large \citep{zhang2019algorithmic}, but this large batch regime is primarily of interest to large organizations which can parallelize training over multiple devices. Second, if we wanted to find algorithms that outperform SGD with small batches and finite compute budgets, the most promising methods would be those that reduce the variance of stochastic gradients \cite{roux2012stochastic}. These algorithms converge significantly faster on convex losses. However, our work confirms that gradient noise has a generalization benefit early in training which leads to higher test accuracies. This may explain why it is difficult to design optimization algorithms for the noise dominated regime that perform well on the test set.



\section*{Acknowledgements}

We thank Brendan O'Donoghue, Andriy Mnih, Chris Maddison, James Martens, Navid Azizan, Tom Goldstein, Razvan Pascanu, Esme Sutherland and Yee Whye Teh for various discussions that have helped improve the paper.

\bibliography{refs}
\bibliographystyle{icml2020}

\newpage
\appendix

\section{Deriving the linear scaling rule for small batch sizes}
\label{app:sde_derive}

In section \ref{sec:two-regimes} of the main text, we applied the central limit theorem to approximate a single SGD step by,
\begin{equation}
    \Delta \omega_i = \left( \omega_{i+1} - \omega_i \right) \approx - \epsilon \frac{dC}{d\omega}\Big|_{\omega = \omega_i} + \sqrt{\epsilon T} \nu_i .
    \label{eq:app_sde}
\end{equation}
The temperature $T = \epsilon/B$, $\mathbb{E}\left(\nu_i \right)  = 0$ and $\mathbb{E} \left( \nu_i \nu_j^\top \right) = F(\omega_i) \delta_{ij}$, where $F(\omega)$ is the empirical Fisher information matrix and $\delta_{ij}$ is the dirac delta function. Equation \ref{eq:app_sde} holds so long as the gradient of each training example is an independent and uncorrelated sample from an underlying short tailed distribution. Additionally, it assumes that the training set size $N \gg B$ and the batch size $B \gg 1$. To derive the linear scaling rule, we consider the total change in the parameters over $n$ consecutive SGD parameter updates,
\begin{equation}
    \Delta \omega_i' = \sum_{j=0}^{n-1} \Delta \omega_{i+j} \approx - \epsilon \left( \sum_{j=0}^{n-1} \frac{dC}{d\omega}\Big|_{\omega = \omega_{i+j}} \right) + \sqrt{n \epsilon T} \xi_i.
    \label{eq:n_updates}
\end{equation}
The noise $\xi_i = (1/\sqrt{n}) \sum_{j=0}^{n-1} \nu_{i+j}$. When the product of the number of steps $n$ and the  learning rate $\epsilon$ is much smaller than the critical learning rate, $n \epsilon \ll \epsilon_{crit}$, the parameters do not move far enough for the gradients to significantly change, and therefore for all $\{j,j'\}$ greater than $0$ and less than $n$,
\begin{eqnarray}
\frac{dC}{d\omega}\Big|_{\omega = \omega_{i+j}} &\approx& \frac{dC}{d\omega}\Big|_{\omega = \omega_{i} } \label{eq:grads_no_move} \\
\mathbb{E} \left( \nu_{i+j} \nu_{i+j'} \right) &\approx& F(\omega_i) \delta_{jj'} \label{eq:cov}
\end{eqnarray}
Using equation \ref{eq:grads_no_move}, we can rewrite equation \ref{eq:n_updates} as
\begin{equation}
    \Delta \omega_i' \approx - n\epsilon \frac{dC}{d\omega}\Big|_{\omega = \omega_i} + \sqrt{n \epsilon T} \xi_i.
    \label{eq:sde_n_final}
\end{equation}
Equation \ref{eq:cov} implies that $\mathbb{E} \left( \xi_i \right) = 0$ and $\mathbb{E} \left( \xi_i \xi_i^\top \right) \approx F(\omega_i)$. We therefore conclude that $\xi$ and $\nu$ are both Gaussian random variables from the same distribution. Comparing equation \ref{eq:app_sde} and equation \ref{eq:sde_n_final}, we conclude that $n$ SGD updates at temperature $T$ with learning rate $\epsilon$ is equivalent to a single SGD step at temperature $T$ with learning rate $n \epsilon$. Since the temperature $T = \epsilon/B$, this implies that when $\epsilon \ll \epsilon_{crit}$, then simultaneously doubling both the learning rate and the batch size should draw samples from the same distribution over parameters after the same number of training epochs.

This prediction is known as the linear scaling rule \citep{krizhevsky2014one, goyal2017accurate, mandt2017stochastic, smith2017bayesian, jastrzkebski2017three, chaudhari2018stochastic, mccandlish2018empirical, shallue2018measuring}. Since this linear scaling rule assumes that $\epsilon \ll \epsilon_{crit}$, it usually holds when the batch size is small, which appears to contradict the assumption $B \gg 1$ above. Crucially however, the distribution of $\nu_i$ does not matter in practice, since our dynamics is governed by the combined influence of noise over multiple consecutive updates, $\xi_i = (1/\sqrt{n}) \sum_{j=0}^{n-1} \nu_{i+j}$.

In other words, we do not require that equation \ref{eq:app_sde} is an accurate model of an single SGD step, we only require that equation \ref{eq:sde_n_final} is an accurate model of $n$ SGD steps. We therefore conclude that $\nu_i$ does not need to be Gaussian, we only require that $\xi_i$ is Gaussian. The central limit theorem predicts that, if $\nu_i$ is an independent random sample from a short-tailed distribution, $\xi_i$ will be Gaussian if $N \gg 1$, $nB \gg 1$ and $nB \ll N$. If $\epsilon \ll \epsilon_{crit}$, then we can choose $1 \ll n \ll N$, and discard the assumption $B \gg 1$.

\section{Additional experimental details}
\label{app:add_exp_details}

In this section, we provide additional details about the experimental setup and models considered in our study.

\subsection{Our learning rate decay schedule}
\label{app:learning_rate_schedule}

\begin{figure}[ht]
\centering
\includegraphics[height=2.8cm]{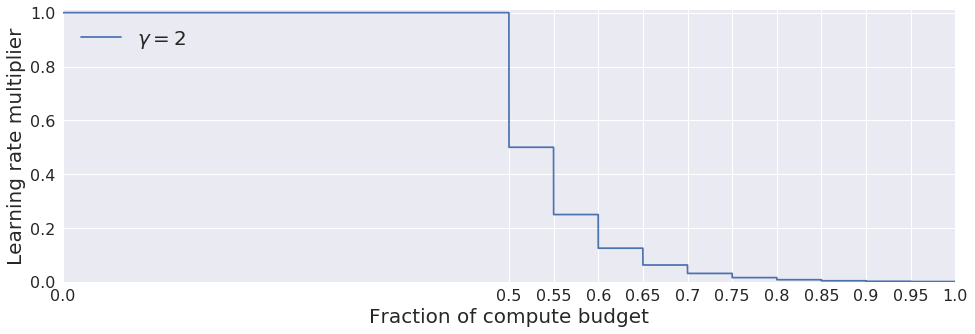}
\caption{Our default learning rate schedule with $\gamma = 2$.}
\label{fig:lr_schedule}
\end{figure}

We illustrate our default learning rate decay schedule in figure \ref{fig:lr_schedule}. As specified in the main text, if the epoch budget is $N_{epochs}$, we hold the learning rate constant for $N_{epochs}/2$, before decaying the learning rate by a factor of $\gamma$ every $N_{epochs}/20$. Unless specified otherwise, $\gamma = 2$. Note that in our constant step experiments, the epoch budget is proportional to the batch size, which ensures that all batch sizes decay the learning rate after the same number of steps.

\subsection{Additional models used}

\begin{figure*}[t]
\centering
\subfigure[Batch size 64]{\includegraphics[height=2.8cm]{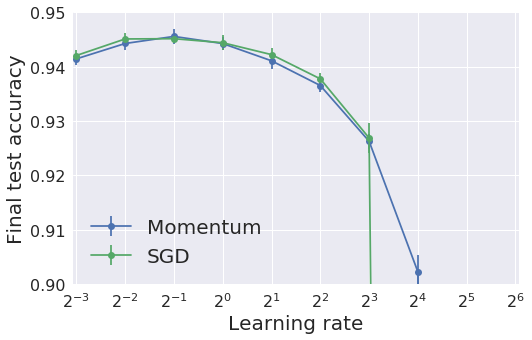}}
\subfigure[Batch size 1024]{\includegraphics[height=2.8cm]{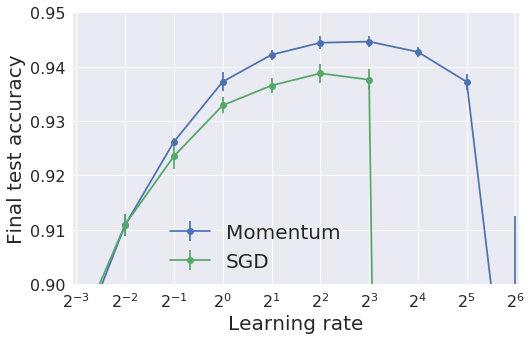}}
\subfigure[Batch size 64]{\includegraphics[height=2.8cm]{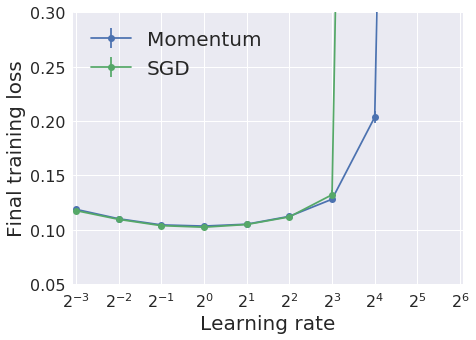}}
\subfigure[Batch size 1024]{\includegraphics[height=2.8cm]{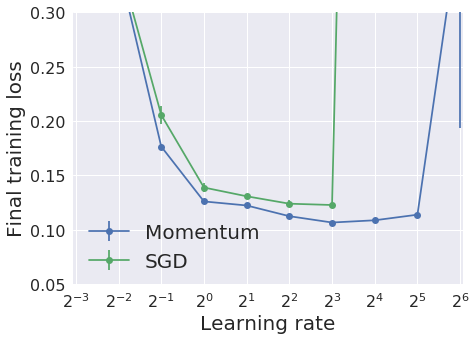}}
\caption{A 16-4 Wide-ResNet, trained with batch normalization on CIFAR-10 for 200 epochs. For completeness, we provide the performance at a range of learning rates for two batch sizes, 64 and 1024 (the performance for a range of batch sizes at the optimal learning rate is shown in figure \ref{fig:two_regimes_constant_epoch_wrn}). The smaller batch size is in the noise dominated regime, while the larger batch size is in the curvature dominated regime. We provide the final test accuracies in figures a and b, and the final training losses at in figures c and d. SGD and SGD with Momentum always achieve similar final performance in the small learning rate limit, while SGD performs poorly when the learning rate is large. When the batch size is small, the optimal learning rate is also small, and so both methods have similar optimal test accuracy/training loss. When the batch size is large, the optimal learning rate is large, and SGD with Momentum performs better.}
\label{fig:mom_learning_rates}
\end{figure*}

\begin{figure*}[t]
\centering
\subfigure[]{\includegraphics[height=3.9cm]{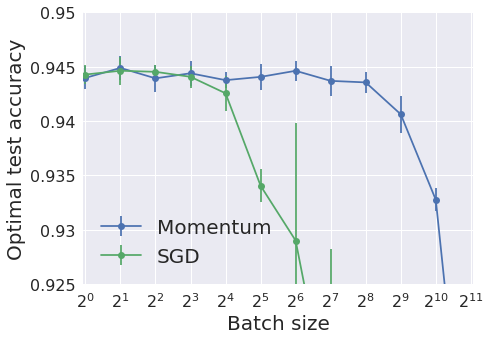}\label{fig:without_BN_constant_epochs_a}}
\subfigure[]{\includegraphics[height=3.9cm]{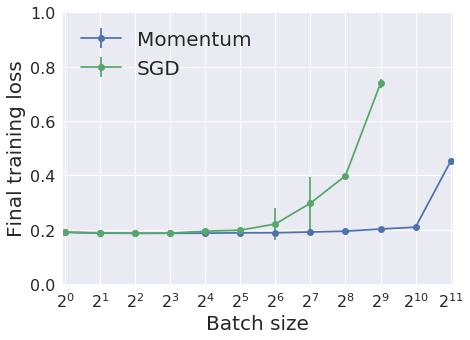}\label{fig:without_BN_constant_epochs_b}}
\subfigure[]{\includegraphics[height=3.9cm]{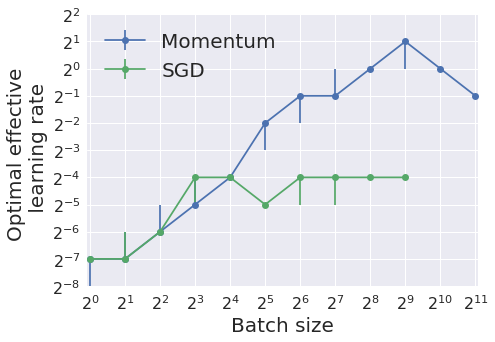}\label{fig:without_BN_constant_epochs_c}}

\caption{A 16-4 Wide-ResNet, trained without batch normalization using Regularized SkipInit \citep{de2020batch} on CIFAR-10 for 200 epochs. We report the performance of SGD and SGD with Momentum. We perform a grid search to identify the optimal learning rate which maximizes the test accuracy, and report the mean performance of the best 12 of 15 runs. a) The test accuracy is independent of batch size when the batch size is small, but begins to fall when the batch size is sufficiently large ($B \gtrsim 8$ for SGD and $B \gtrsim 256$ for SGD with Momentum. b) The training loss at the optimal effective learning rate is independent of batch size when the batch size is small, but rises rapidly when the batch size is sufficiently large. c) The optimal effective learning rate is proportional to batch size when the batch size is small for both both SGD and SGD with Momentum, while it is independent of batch size when the batch size is sufficiently large. 
\label{fig:without_BN_constant_epochs}}
\end{figure*}

\begin{table*}[t]
\centering
\caption{ResNet-50, trained on ImageNet for 90 epochs. We follow the implementation of \citet{goyal2017accurate}, however we introduce our modified learning rate schedule defined in appendix \ref{app:learning_rate_schedule}. We perform a grid search to identify the optimal effective learning rate and report the performance of a single training run. The test accuracies achieved by SGD and Momentum are equal when the batch size is small, but Momentum outperforms SGD when the batch size is large. For SGD with Momentum, the optimal effective learning rate is proportional to batch size for all batch sizes considered, while this linear scaling rule breaks at large batch sizes for SGD.\\}
\begin{tabular}{c|c|c|c|c}
\textbf{} & \textbf{Batch size} & \textbf{Optimal test accuracy (\%)} & \textbf{Training loss} & \textbf{Optimal effective learning rate} \\ \hline
          & 256                 & 77.0                                & 2.25                   & 1.0                            \\
SGD       & 1024                & 76.7                                & 2.25                   & 4.0                            \\
          & 4096                &           76.1                          &  2.30                      &    8.0                            \\ \hline
          & 256                 & 77.0                                & 2.25                   & 1.0                            \\
Momentum  & 1024                & 76.8                                & 2.25                   & 4.0                            \\
          & 4096                & 76.8                                    & 2.25                       & 16.0                               
\end{tabular}
\label{table:imagenet_constant_epoch}
\end{table*}

In addition to the 16-4 Wide-ResNet model on CIFAR-10 presented in the main paper \citep{zagoruyko2016wide}, we provide additional experiments in the appendix using ResNet-50 on ImageNet \citep{he2016deep}, 28-10 Wide-ResNet on CIFAR-100 \citep{zagoruyko2016wide}, LSTMs on Penn TreeBank \citep{zaremba2014recurrent} and autoencoders on MNIST \citep{sutskever2013importance}.

\textbf{ResNet50 on ImageNet: } We follow the modified ResNet-50 implementation of \citet{goyal2017accurate} for training on ImageNet, and we use our default learning rate schedule without learning rate warmup (see appendix \ref{app:learning_rate_schedule}). Due to the large compute budget required for these models, we train a single model for each batch size/learning rate pair.

\textbf{28-10 Wide-ResNet on CIFAR100: } We train 28-10 Wide-ResNets on CIFAR-100 \citep{zagoruyko2016wide}. We use our default learning rate schedule, as described in appendix \ref{app:learning_rate_schedule}, which reaches the same test set accuracy as is reported by \citet{zagoruyko2016wide}.

\textbf{LSTM on Penn TreeBank: } We train a word-level LSTM language model on the Penn TreeBank dataset (PTB), following the implementation described in \citet{zaremba2014recurrent}. The LSTM model used has two layers with 650 units per layer. The parameters are initialized uniformly in $[-0.05, 0.05]$. We apply gradient clipping at $5$, as well as dropout with probability $0.5$ on the non-recurrent connections. We train the LSTM using an unroll step of 35, and use the learning rate decay schedule described in appendix \ref{app:learning_rate_schedule}. As with the other models tested in this paper, this learning rate schedule reaches the same test perplexity performance as the original schedules reported in \cite{zaremba2014recurrent}.

\begin{figure*}[t]
\centering
\subfigure[]{\includegraphics[height=4cm]{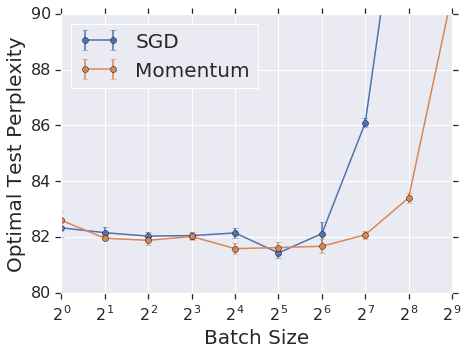}}
\subfigure[]{\includegraphics[height=4cm]{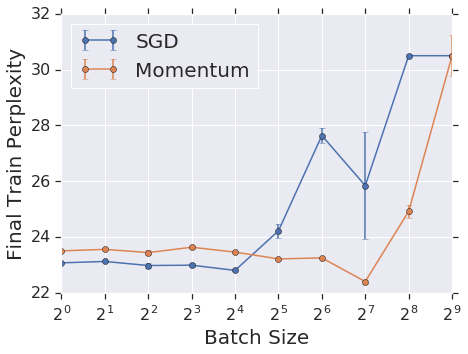}}
\subfigure[]{\includegraphics[height=4cm]{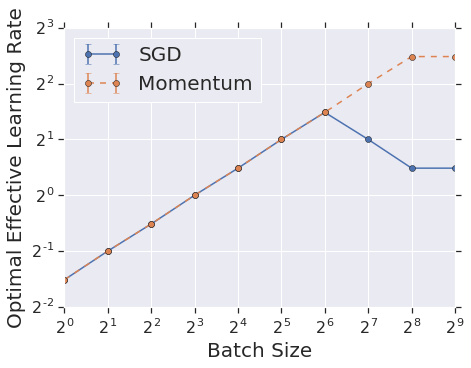}}
\caption{A word-level LSTM language model trained on PTB for 40 epochs. We report the performance of SGD and SGD with Momentum. We perform a grid search to identify the optimal learning rate which maximizes the test set perplexity, and report the mean performance of the best 5 of 7 runs.  a) The test set perplexity of SGD with Momentum is independent of batch size when the batch size is small, but begins to rise when the batch size exceeds 128. The test set perplexity of vanilla SGD starts rising for batch sizes exceeding 64. b) We see similar phenomena on the training set perplexity.  c) Surprisingly, the optimal effective learning rate is proportional to \emph{square root} of the batch size when the batch size is small, while it levels off for larger batch sizes. We note that the gradients of consecutive minibatches in a language model are not independent, which violates the assumptions required to derive the linear scaling rule.}
\label{fig:lstm_constant_epoch}
\end{figure*}

\begin{figure*}[t]
\centering
\subfigure[]{\includegraphics[height=4cm]{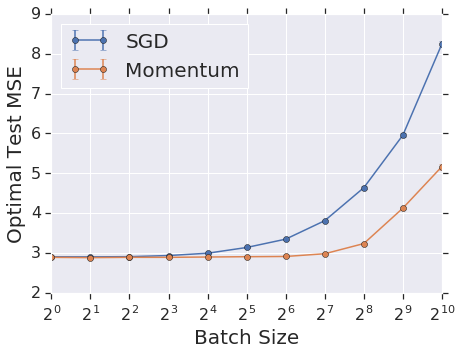}}
\subfigure[]{\includegraphics[height=4cm]{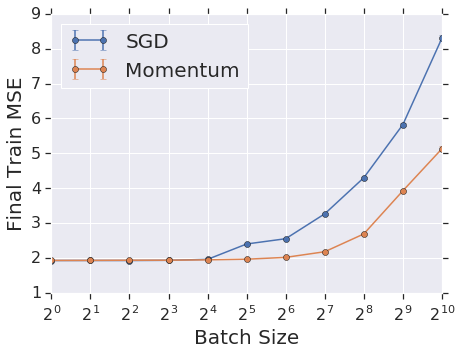}}
\subfigure[]{\includegraphics[height=4cm]{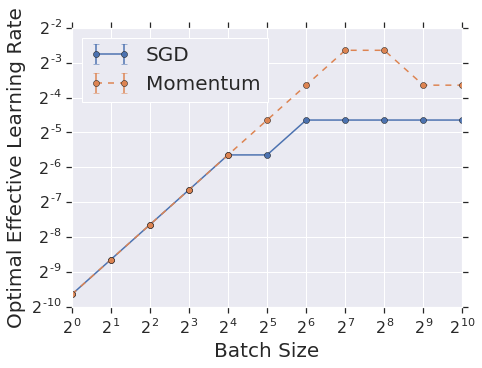}}
\caption{A fully connected autoencoder, trained on MNIST for 200 epochs. We report the performance of SGD and SGD w/ Momentum. We perform a grid search to identify the optimal learning rate which maximizes the mean-squared error (MSE) on the test set, and report the mean performance of the best 5 of 7 runs.  a) The test MSE of SGD w/ Momentum is initially independent of batch size, but it begins to rise when the batch size exceeds 128. The test MSE of vanilla SGD starts rising for batch sizes exceeding 16. b) We see similar phenomena on the training set MSE.  c) The optimal effective learning rate is proportional to batch size when the batch size is small for both vanilla SGD and SGD w/ Momentum, while it becomes independent of batch size for larger batch sizes. The optimal effective learning rate in the curvature dominated regime is larger for SGD w/ Momentum.}
\label{fig:ae_constant_epoch}
\end{figure*}

\textbf{Autoencoder on MNIST: } We train a fully-connected autoencoder on MNIST. Our network architecture is described by the sequence of layer widths $\{784, 1000, 500, 250, 30, 250, 500, 1000, 784\}$, where $784$ denotes the input and output dimensions. For more details on this architecture, we refer to \citet{sutskever2013importance}. This model has often been used as an optimization benchmark \citep{sutskever2013importance, kidambi2018insufficiency}. The L2 regularization parameter was set at $10^{-5}$. We use our default learning rate schedule, as described in appendix \ref{app:learning_rate_schedule}.

\section{Additional results under constant epoch budgets}
\label{app:additional_results_constant_epoch}

In this section, we provide additional experimental results to verify the existence of two regimes of SGD under a constant epoch budget. In all cases, we observe a transition from a small batch regime, where the learning rate increases with the batch size and SGD with Momentum does not outperform SGD, to a large batch regime, where the learning rate is independent of the batch size and SGD with Momentum outperforms SGD. Under a constant epoch budget, both the training loss and the optimal test accuracy are independent of batch size in the noise dominated regime, but begin to degrade when one enters the curvature dominated regime.

\subsection{Learning rate sweep with batch normalization for two batch sizes on CIFAR-10 Wide-ResNet model}
\label{app:lr_sweep_wrn}

In figure \ref{fig:mom_learning_rates}, we provide additional results with the 16-4 Wide-ResNet, trained with batch normalization on CIFAR-10 for 200 epochs. Here we provide the final test set accuracies and the final training set losses for a full learning rate sweep at two batch sizes, 64 and 1024. From figure \ref{fig:mom_learning_rates}, we see that SGD and Momentum always achieve similar final performance in the small learning rate limit. This confirms previous theoretical work showing the equivalence of SGD and Momentum in the small learning rate limit when the momentum parameter is kept fixed \citep{orr1994momentum, qian1999momentum, yuan2016influence}. Meanwhile, SGD performs poorly compared to SGD with Momentum when the learning rate is large. When the batch size is small, the optimal learning rates for both methods are also small, and so the two methods have the same optimal test accuracy. However when the batch size is large, the optimal learning rate is large, and consequently SGD with Momentum outperforms vanilla SGD. These results are entirely consistent with the two regimes of SGD discussed in section \ref{sec:two-regimes}.

\subsection{Results without batch normalization on CIFAR-10 Wide-ResNet}
\label{app:without_BN_constant_epoch}

In figure \ref{fig:without_BN_constant_epochs} we present results when training our 16-4 Wide-ResNet \citep{zagoruyko2016wide}. We follow the same setup and learning rate schedule described in section \ref{sec:exp_setup}, and we train for 200 epochs. However we remove batch normalization, and introduce the Regularized SkipInit initialization scheme proposed by \citet{de2020batch}. This initialization scheme enables the training of very deep networks, and it reduces the gap in test accuracy between networks trained with and without batch normalization.

We observe remarkably similar trends to those observed in section \ref{sec:constant-epoch} of the main text, although the critical learning rate, beyond which the optimal learning rate of SGD is independent of batch size, is significantly smaller when batch normalization is not used. The performance of SGD on both the training and the test set is independent of batch size for very small batch sizes $B \lesssim 8$, while the performance of SGD with Momentum is constant for batch sizes $B \lesssim 256$. Above these thresholds, the performance of both methods degrades rapidly. These observations are explained by the optimal effective learning rates in figure \ref{fig:without_BN_constant_epochs_c}. SGD with Momentum has a significantly larger maximum stable learning rate, enabling it to scale to larger batch sizes.

\subsection{Results from additional models}
\label{app:additional_models_constant_epoch}

\begin{table*}[t]
\centering
\caption{The optimal test accuracy and final training loss for a range of batch sizes under a constant step budget. For each batch size, we train a 16-4 Wide-ResNet without batch normalization on CIFAR-10 using Regularized SkipInit \citep{de2020batch} for 156,250 updates. We perform a grid search to identify the optimal learning rate which maximizes the test accuracy, and we provide the average performance of the best 12 out of 15 training runs. The final test accuracy falls for very large batches. We note that, although the final training loss rises slightly at batch size 2048, the training loss remains lower than that achieved at batch size 128 (for which the test accuracy was maximized). \\}
\begin{tabular}{cccc}
\multicolumn{1}{c|}{\textbf{Batch size}} & \multicolumn{1}{c|}{\textbf{Optimal test accuracy}} & \multicolumn{1}{c|}{\textbf{Final training loss}} & \textbf{Optimal effective learning rate} \\ \hline
\multicolumn{1}{c|}{16}                  & \multicolumn{1}{c|}{$92.6 \pm 0.2$}                       & \multicolumn{1}{c|}{$0.349 \pm 0.004$}                        & $2^{-4} \,\,\, (2^{-4} \textrm{ to } 2^{-2})$     \\
\multicolumn{1}{c|}{32}                  & \multicolumn{1}{c|}{$93.9 \pm 0.1$}                       & \multicolumn{1}{c|}{$0.269 \pm 0.004$}                        & $2^{-3} \,\,\, (2^{-3} \textrm{ to } 2^{-2})$     \\ 
\multicolumn{1}{c|}{64}                  & \multicolumn{1}{c|}{$94.4 \pm 0.1$}                       & \multicolumn{1}{c|}{$0.192 \pm 0.002$}                        & $2^{-2} \,\,\, (2^{-2} \textrm{ to } 2^{-1})$                       \\
\multicolumn{1}{c|}{128}                  & \multicolumn{1}{c|}{$94.6 \pm 0.1$}                       & \multicolumn{1}{c|}{$0.122 \pm 0.000$}                        & $2^{-1} \,\,\, (2^{-1} \textrm{ to } 2^{-1})$                       \\
\multicolumn{1}{c|}{256}                 & \multicolumn{1}{c|}{$94.4 \pm 0.1$}                       & \multicolumn{1}{c|}{$0.071 \pm 0.001$}                         & $2^{-1} \,\,\, (2^{-1} \textrm{ to } 2^{0})$      \\
\multicolumn{1}{c|}{512}                 & \multicolumn{1}{c|}{$94.1 \pm 0.1$}                       & \multicolumn{1}{c|}{$0.043 \pm 0.000$}                         & $2^{-0} \,\,\, (2^{-1} \textrm{ to } 2^{1})$      \\
\multicolumn{1}{c|}{1024}                & \multicolumn{1}{c|}{$93.8 \pm 0.1$}                       & \multicolumn{1}{c|}{$0.028 \pm 0.000$}                         & $2^{-1}  \,\,\, (2^{-1} \textrm{ to } 2^{0})$                          \\
\multicolumn{1}{c|}{2048}                & \multicolumn{1}{c|}{$93.0 \pm 0.6$}                       & \multicolumn{1}{c|}{$0.054 \pm 0.050$}                         & $2^{-1}  \,\,\, (2^{-1} \textrm{ to } 2^{0})$                          \\

\end{tabular}
\label{table:constant_step_without_bn}
\end{table*}


\begin{table*}[]
\centering
\caption{The optimal test accuracy and final training loss for a range of batch sizes under a constant step budget. For each batch size, we train a 28-10 Wide-ResNet with batch normalization on CIFAR-100 for 9765 updates. We perform a grid search to identify the optimal learning rate which maximizes the test accuracy, and we provide the average performance of the best 12 out of 15 training runs. The test accuracy initially rises as the batch size rises, but it falls for very large batches. The training loss also rises at the largest batch size considered of 16384 but remains comparable to that observed at batch size 2048 (for which the test accuracy was maximized). \\}
\begin{tabular}{c|c|c|c}
\textbf{Batch size} & \textbf{Optimal test accuracy (\%)} & \textbf{Final training loss} & \textbf{Optimal effective learning rate} \\ \hline
256                 & $78.9 \pm 0.1$                      & $0.609 \pm 0.004$            & $2^2$ ($2^2$ to $2^2$)                   \\
512                 & $79.9 \pm 0.2$                      & $0.462 \pm 0.008$            & $2^3$ ($2^2$ to $2^3$)                   \\
1024                & $80.1 \pm 0.2$                      & $0.274 \pm 0.003$            & $2^4$ ($2^3$ to $2^4$)                   \\
2048                & $80.2 \pm 0.2$                      & $0.132 \pm 0.002$            & $2^4$ ($2^4$ to $2^4$)                   \\
4096                & $79.6 \pm 0.2$                      & $0.073 \pm 0.001$            & $2^5$ ($2^5$ to $2^5$)                   \\
8192                & $78.1 \pm 0.4$                      & $0.045 \pm 0.001$            & $2^5$ ($2^1$ to $2^5$)     \\
16384                & $72.2 \pm 0.2$                      & $0.156 \pm 0.036$            & $2^1$ ($2^1$ to $2^2$)  
\end{tabular}
\label{table:constant_step_cifar100_with_bn}
\end{table*}

\begin{table*}[t]
\centering
\caption{The optimal test accuracy and final training loss for a range of batch sizes under a constant step budget. For each batch size, we train a 28-10 Wide-ResNet on CIFAR-100 without batch normalization using Regularized SkipInit for 156,250 updates. We perform a grid search to identify the optimal learning rate which maximizes the test accuracy, and we provide the average performance of the best 12 out of 15 training runs. The final test accuracy falls for very large batches, while surprisingly the training loss also rises slightly. \\
}
\begin{tabular}{cccc}
\multicolumn{1}{c|}{\textbf{Batch size}} & \multicolumn{1}{c|}{\textbf{Optimal test accuracy}} & \multicolumn{1}{c|}{\textbf{Final training loss}} & \textbf{Optimal effective learning rate} \\ \hline
\multicolumn{1}{c|}{32}                  & \multicolumn{1}{c|}{$77.9 \pm 0.2$}                       & \multicolumn{1}{c|}{$0.0512 \pm 0.0611$}                        & $0.025 \,\,\, (0.025 \textrm{ to } 0.050)$                       \\
\multicolumn{1}{c|}{64}                  & \multicolumn{1}{c|}{$79.0 \pm 0.2$}                       & \multicolumn{1}{c|}{$0.0141 \pm 0.0281$}                        & $0.050 \,\,\, (0.050 \textrm{ to } 0.050)$                       \\
\multicolumn{1}{c|}{128}                  & \multicolumn{1}{c|}{$79.4 \pm 0.1$}                       & \multicolumn{1}{c|}{$0.0011 \pm 0.0004$}                        & $0.100 \,\,\, (0.100 \textrm{ to } 0.100)$                       \\
\multicolumn{1}{c|}{256}                  & \multicolumn{1}{c|}{$78.9 \pm 0.1$}                       & \multicolumn{1}{c|}{$0.0191 \pm 0.0221$}                        & $0.200 \,\,\, (0.200 \textrm{ to } 0.200)$                       \\
\multicolumn{1}{c|}{512}                  & \multicolumn{1}{c|}{$77.7 \pm 0.1$}                       & \multicolumn{1}{c|}{$0.0022 \pm 0.0035$}                        & $0.200 \,\,\, (0.200 \textrm{ to } 0.200)$                       \\
\multicolumn{1}{c|}{1024}                  & \multicolumn{1}{c|}{$76.0 \pm 0.1$}                       & \multicolumn{1}{c|}{$0.0009 \pm 0.0012$}                        & $0.200 \,\,\, (0.200 \textrm{ to } 0.200)$                       \\
\multicolumn{1}{c|}{2048}                  & \multicolumn{1}{c|}{$74.2 \pm 0.3$}                       & \multicolumn{1}{c|}{$0.0029 \pm 0.0025$}                        & $0.200 \,\,\, (0.200 \textrm{ to } 0.200)$                       \\
\end{tabular}
\label{table:constant_step_cifar100_without_bn}
\end{table*}

\begin{table*}[t]
\centering
\caption{The optimal test set perplexity and final training set perplexity for a range of batch sizes under a constant step budget. For each batch size, we train a word-level LSTM on PTB for 16560 updates. We perform a grid search to identify the optimal learning rate which maximizes the test set perplexity, and we provide the average performance of the best 5 out of 7 training runs. The optimal test perplexity increases as the batch size rises, while the training perplexity falls. \\
}
\begin{tabular}{cccc}
\multicolumn{1}{c|}{\textbf{Batch size}} & \multicolumn{1}{c|}{\textbf{Optimal test perplexity}} & \multicolumn{1}{c|}{\textbf{Final training perplexity}} & \textbf{Optimal effective learning rate} \\ \hline
\multicolumn{1}{c|}{16}                  & \multicolumn{1}{c|}{$88.06 \pm 0.26$}                       & \multicolumn{1}{c|}{$35.85 \pm 0.06$}                        & $1.4 \,\,\, (1.4 \textrm{ to } 1.4)$                       \\
\multicolumn{1}{c|}{32}                  & \multicolumn{1}{c|}{$82.50 \pm 0.13$}                       & \multicolumn{1}{c|}{$28.99 \pm 0.03$}                        & $2.0 \,\,\, (2.0 \textrm{ to } 2.0)$                       \\
\multicolumn{1}{c|}{64}                  & \multicolumn{1}{c|}{$81.67 \pm 0.26$}                       & \multicolumn{1}{c|}{$23.25 \pm 0.05$}                        & $2.8 \,\,\, (2.8 \textrm{ to } 2.8)$                       \\
\multicolumn{1}{c|}{128}                & \multicolumn{1}{c|}{$86.04 \pm 0.49$}                       & \multicolumn{1}{c|}{$21.16 \pm 0.06$}                         & $5.6 \,\,\, (5.6 \textrm{ to } 5.6)$                          \\
\multicolumn{1}{c|}{256}                 & \multicolumn{1}{c|}{$92.19 \pm 0.26$}                       & \multicolumn{1}{c|}{$15.74 \pm 0.03$}                         & $5.6 \,\,\, (5.6 \textrm{ to } 5.6)$                         \\
\end{tabular}
\label{table:lstm_constant_step}
\end{table*}

\begin{table*}[t]
\centering
\caption{The optimal test set MSE and final training set MSE for a range of batch sizes under a constant step budget. For each batch size, we train a fully connected autoencoder on MNIST for 156,250 updates. We perform a grid search to identify the optimal learning rate which maximizes the test set MSE, and we provide the average performance of the best 5 out of 7 training runs. The final test MSE falls for large batch sizes, although this effect is rather weak in this model. \\
}
\begin{tabular}{cccc}
\multicolumn{1}{c|}{\textbf{Batch size}} & \multicolumn{1}{c|}{\textbf{Optimal test set MSE}} & \multicolumn{1}{c|}{\textbf{Final training set MSE}} & \textbf{Optimal effective learning rate} \\ \hline
\multicolumn{1}{c|}{64}                  & \multicolumn{1}{c|}{$2.91 \pm 0.01$}                       & \multicolumn{1}{c|}{$2.017 \pm 0.003$}                        & $0.08 \,\,\, (0.08 \textrm{ to } 0.08)$                       \\
\multicolumn{1}{c|}{256}                 & \multicolumn{1}{c|}{$2.95 \pm 0.01$}                       & \multicolumn{1}{c|}{$2.010 \pm 0.005$}                         & $0.08 \,\,\, (0.08 \textrm{ to } 0.08)$                         \\
\multicolumn{1}{c|}{1024}                & \multicolumn{1}{c|}{$2.96 \pm 0.01$}                       & \multicolumn{1}{c|}{$2.005 \pm 0.011$}                         & $0.08  \,\,\, (0.08 \textrm{ to } 0.08)$                          \\
\multicolumn{1}{c|}{4096}                & \multicolumn{1}{c|}{$2.98 \pm 0.01$}                       & \multicolumn{1}{c|}{$2.018 \pm 0.008$}                         & $0.08  \,\,\, (0.08 \textrm{ to } 0.08)$                          \\
\end{tabular}
\label{table:ae_constant_step}
\end{table*}

\begin{figure*}[t]
\centering
\subfigure[]{\includegraphics[height=4cm]{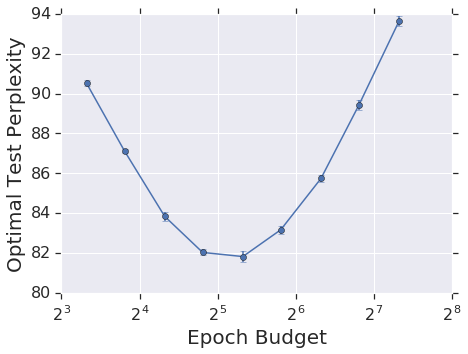}}
\subfigure[]{\includegraphics[height=4cm]{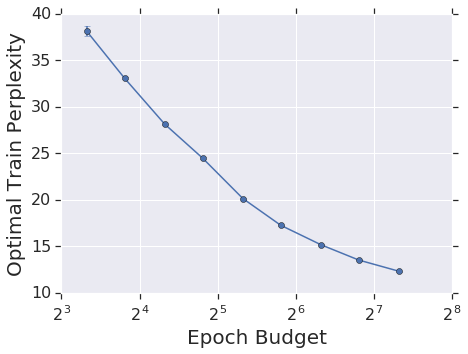}}
\subfigure[]{\includegraphics[height=4cm]{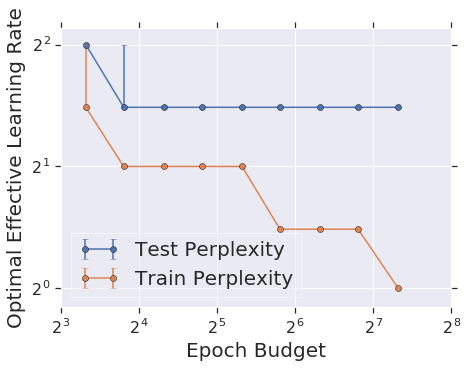}}
\caption{The performance of a word-level LSTM language model trained on the Penn TreeBank dataset using SGD with Momentum and a batch size of 64 at a range of epoch budgets. We identify both the optimal effective learning rate which minimizes the test set perplexity and the optimal effective learning rate which minimizes the training set perplexity, and we present the mean performance of the best 5 out of 7 runs. a) Initially the test set perplexity falls as the epoch budget increases, however it begins to rise beyond $56$ training epochs. b) The training set perplexity falls monotonically as the epoch budget rises. c) The learning rate that minimizes the training set perplexity falls as the epoch budget rises, while the learning rate that minimizes the test set perplexity only varies by a factor of 2 when the epoch budget rises over two orders of magnitude.}
\label{fig:lstm_lr_vs_compute_budget}
\end{figure*}

\begin{figure*}[t]
\centering
\subfigure[]{\includegraphics[height=3.1cm]{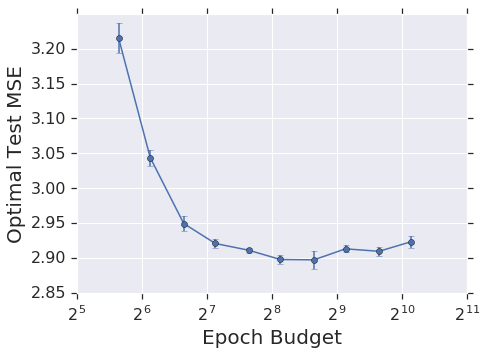}\label{fig:ae_lr_vs_compute_budget_a}}
\subfigure[]{\includegraphics[height=3.1cm]{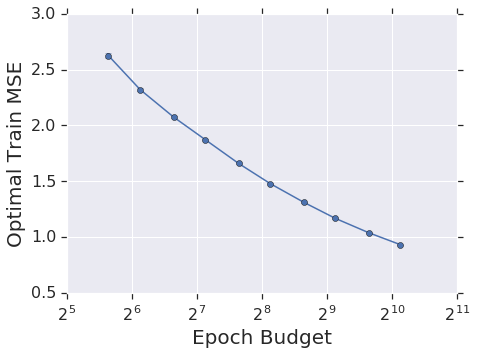}\label{fig:ae_lr_vs_compute_budget_b}}
\subfigure[]{\includegraphics[height=3.1cm]{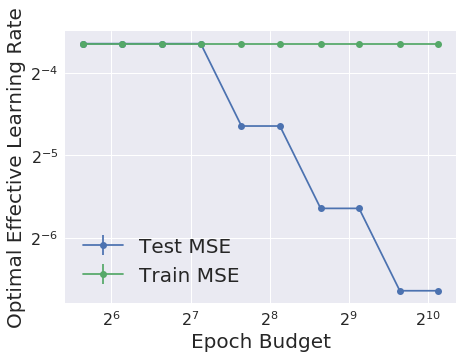}\label{fig:ae_lr_vs_compute_budget_c}}
\subfigure[]{\includegraphics[height=3.1cm]{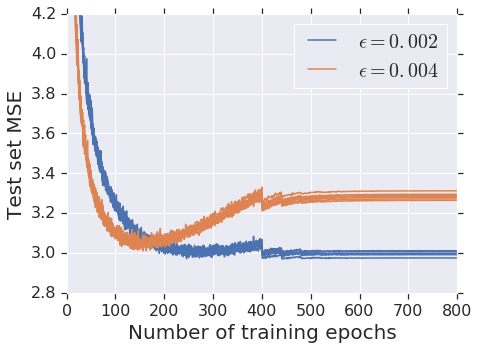}\label{fig:ae_lr_vs_compute_budget_d}}
\caption{The performance of a fully connected autoencoder on MNIST using SGD with Momentum and a batch size of 32 at a range of epoch budgets. We identify both the optimal effective learning rate which minimizes the test set MSE and the optimal effective learning rate which minimizes the training set MSE, and we present the mean performance of the best 5 out of 7 runs. a) Initially the test set MSE falls as the epoch budget increases, before rising slightly for large epoch budgets. b) The train set MSE falls monotonically as the the epoch budget rises. c) The learning rate that minimizes the test set MSE decreases, while the learning rate that minimizes the train set MSE remains constant as the epoch budget rises. This is contrary to what we observe in figures \ref{fig:cifar_lr_vs_epoch_budget} and  \ref{fig:lstm_lr_vs_compute_budget}. The reason for this is apparent from figure d), where we plot the test set MSE during training for all 7 runs for an epoch budget of 800 for learning rate $\epsilon = 0.004$ and $\epsilon = 0.002$. We notice that for a larger learning rate, the model overfits on the training set faster, causing the test set MSE to rise by the time of the first learning rate drop at 400 epochs. This suggests that early stopping has more influence on the final test performance in this architecture than stochastic gradient noise.}
\label{fig:ae_lr_vs_compute_budget}
\end{figure*}

In table \ref{table:imagenet_constant_epoch}, we provide results for ResNet-50 trained on ImageNet for 90 epochs at a small range of batch sizes. SGD with and without Momentum achieve similar test accuracies when the batch size is small, but SGD with Momentum outperforms SGD without Momentum when the batch size is large. The optimal effective learning rate is proportional to batch size for all batch sizes considered when using SGD with Momentum, but not when using vanilla SGD.

In figure \ref{fig:lstm_constant_epoch}, we present results for the LSTM on PTB trained for 40 epochs. Once again, we see that SGD and SGD with Momentum have similar performance for small batch sizes. Performance for SGD starts degrading for batch sizes exceeding 64, whereas performance for SGD with Momentum starts degrading for batch sizes exceeding 128. However, unlike our previous experiments, we notice that the optimal learning rate increases as \emph{square root} of the batch size for small batch sizes, before leveling off at a constant value for larger batch sizes. The square root scaling observed here could be due to correlations between consecutive data samples when training the LSTM, which violate the assumptions used to derive the linear scaling rule in section \ref{sec:two-regimes}.

In figure \ref{fig:ae_constant_epoch}, we present results on training the fully-connected autoencoder on MNIST for 200 epochs. 
As before, we notice that for small batch sizes, the performance of both SGD and SGD with Momentum is independent of batch size, while performance begins to degrade when the batch size is large. On this model, the performance of SGD begins to  degrade at much smaller batch sizes than we observed in normalized residual networks, and consequently SGD with Momentum starts outperforming SGD at much smaller batch sizes. This is likely due to the poor conditioning of the model due to the bottleneck structure of its architecture.


\section{Additional results under constant step budgets}
\label{app:additional_results_constant_step}

In this section, we provide additional results studying how the optimal test accuracy depends on the batch size under a constant step budget for a range of models. We train with SGD with Momentum. In each case, we set the number of training steps to be equal to the number of training steps taken by the largest batch size before performance starts degrading under our constant epoch budget experiments.

In table \ref{table:constant_step_without_bn}, we show results for training a 16-4 Wide-ResNet on CIFAR-10 without batch normalization using Regularized SkipInit \citep{de2020batch} for 156250 updates. This corresponds to 200 epochs when the batch size is 64. The final training loss falls as the batch size rises, while the test accuracy drops for large batch sizes. At the largest batch size considered, both the test accuracy and the training loss exhibit a large standard deviation across different training runs (at the optimal learning rate). At this batch size, we note that the variance is much lower at lower learning rates (at which the training loss is also lower), however these smaller learning rates also achieve lower mean test accuracy.

In table \ref{table:constant_step_cifar100_with_bn}, we train a 28-10 Wide-ResNet on CIFAR-100 with batch normalization for 9765 steps at a range of batch sizes (which corresponds to 200 epochs when the batch size is 1024), while in table \ref{table:constant_step_cifar100_without_bn} we train a 28-10 Wide-ResNet on CIFAR-100 without batch normalization (using Regularized SkipInit) for 156,250 steps (which
corresponds to 200 epochs when the batch size is 64). In both cases the test accuracy drops significantly when the batch size is very large. In table \ref{table:constant_step_cifar100_with_bn} the training loss falls as the batch size rises, while in \ref{table:constant_step_cifar100_without_bn} the training loss rises slightly for very large batches.

In table \ref{table:lstm_constant_step}, we train the word-level LSTM on the Penn TreeBank (PTB) dataset \citep{zaremba2014recurrent} for 16560 updates. This corresponds to 40 epochs at batch size 64. We described this model in appendix \ref{app:add_exp_details}, and we train using the learning rate schedule defined in appendix \ref{app:learning_rate_schedule} using SGD with Momentum. The test perplexity increases as the batch size increases, while the training perplexity falls.

In table \ref{table:ae_constant_step}, we train a fully connected auto-encoder on MNIST for 156,250 updates \citep{sutskever2013importance}. This corresponds to 200 epochs when the batch size is 64. We described this model in appendix \ref{app:add_exp_details}, and we train using the learning rate schedule defined in appendix \ref{app:learning_rate_schedule} using SGD with Momentum. The test set MSE increases slightly as the batch size increases, while the training set MSE falls as the batch size rises. Although the training set MSE does appear to rise slightly for a batch size of 4096, we note that the training loss in this case is similar to that achieved with a batch size of 64, while the test set MSE in this case is worse than that at batch size 64.
The optimal effective learning rate is independent of the batch size, suggesting that the learning rate may be close to curvature dominated regime. We note that the benefits of noise appear to be significantly smaller in this architecture than for the Wide-ResNet or LSTM.

\section{Additional results with a fixed batch size and variable epoch budget}
\label{app:additional_results_variable_epoch_budget}

We now provide additional experimental results to accompany those provided in section \ref{sec:implicit-reg}, where we study whether the optimal training temperature is independent of the epoch budget. We use SGD with Momentum with the momentum parameter $m=0.9$ for all our experiments in this section. 

In figure \ref{fig:lstm_lr_vs_compute_budget}, we present results on a word-level LSTM on the PTB dataset for a batch size of 64 and for varying epoch budgets. Note that the original LSTM model in \citet{zaremba2014recurrent} was trained for 39 epochs. The results in figure \ref{fig:lstm_lr_vs_compute_budget} are remarkably similar to those presented in figure \ref{fig:cifar_lr_vs_epoch_budget}. As the epoch budget rises, the test set perplexity first falls but then begins to increase. The training set perplexity falls monotonically as the epoch budget increases. Finally, the optimal learning rate which minimizes the test set perplexity is independent of the epoch budget once this epoch budget is not too small, while the optimal learning rate which minimizes the training set perplexity falls.

In figure \ref{fig:ae_lr_vs_compute_budget}, we present results on a fully connected autoencoder trained on MNIST for a batch size of 32 and for a range of epoch budgets. Note that the autoencoder results presented in section \ref{app:additional_results_constant_epoch} were trained for 200 epochs. Figures \ref{fig:ae_lr_vs_compute_budget_a} and \ref{fig:ae_lr_vs_compute_budget_b} are similar to figures \ref{fig:cifar_lr_vs_epoch_budget_a} and \ref{fig:cifar_lr_vs_epoch_budget_b} in the main text. Initially the test set MSE falls as the epoch budget increases, but then it starts increasing. The training set MSE falls monotonically as the epoch budget rises. In figure \ref{fig:ae_lr_vs_compute_budget_c} however, we notice that the learning rate that minimizes the test set MSE decreases as the epoch budget rises. This is the opposite of what we observed in figures \ref{fig:cifar_lr_vs_epoch_budget} and  \ref{fig:lstm_lr_vs_compute_budget}. To further investigate this, in figure \ref{fig:ae_lr_vs_compute_budget_d} we plot the mean test set MSE during training for an epoch budget of 800 for learning rates $\epsilon = 0.004$ and $\epsilon = 0.002$. We notice that for the larger learning rate $\epsilon = 0.004$, the model overfits faster on the training set, causing the test set MSE to rise by the time of the first learning rate drop at 400 epochs. This is consistently the case for all epoch budgets over 200 epochs. To avoid the test set MSE from rising, the optimal learning rate for the test MSE drops to slow down training sufficiently such that there is no overfitting before the first learning rate decay. Meanwhile the optimal learning rate to minimize the training loss is more or less constant. This suggests that early stopping is particularly important in this architecture and dataset, and that it has more influence on the final test performance than stochastic gradient noise. 

\end{document}